\documentclass{article}
\usepackage[utf8]{inputenc}

\usepackage{amsmath}
\newenvironment{definition}[1]{\par\noindent\underline{Definition:}\space#1}{}
\title{Parametric PDF for Goodness of Fit }
\author{Natan Katz \\  {natan.katz@gmail.com}    \and Uri Utai \\   {uri.itai@gmail.com}  }

\date{October 2022}
\vspace{0.9cm}

\usepackage{csquotes}
\usepackage[utf8]{inputenc}
\usepackage[english]{babel}
\usepackage{subcaption}
\usepackage{graphicx}
 
\graphicspath{{plot_fold/}}
\usepackage[
backend=biber,
style=alphabetic,
sorting=ynt
]{biblatex}
\begin{document}
\maketitle
 
\begin{abstract}
The methods for the goodness of fit in classification problems 
require a prior threshold for determining the confusion matrix.
Nonetheless, this fixed threshold removes information that the model's curves provide, and can be used, for further studies such as risk evaluation and stability analysis.
We present a different framework that allows us to perform this study using a  parametric PDF.  
\end{abstract}

\section{Introduction}
Machine learning (ML) projects have become a leading tool in enormous domains of the computer industry. Their rule is far beyond computational aspects. Indeed, they are a focal point in designing analytical business decisions.
The commercial usage of these models raises new challenges. 
The ML academic research often assumes that : 
\begin{itemize}
\item The data in the database represents well the global
 data distribution.
\item Training methodology aligns with the model's KPI.
\item There are no production-driven drawbacks.
\end{itemize}
Unfortunately, none of these assumptions hold in real-world models.
In addition, cardinal issues that focus on complexity and stability and questions such as \textbf{"what is the efficient way to set a threshold to have both good and stable performance"}  rarely exist in the academy. 
Hence, deploying ML models in the real world requires a methodology that the academy does not provide.    
In the academy, researchers focus mainly on common KPIs such as accuracy and precision. We use these KPIs for other scaling indicators such as Creamer's V, F1-score, AUC \cite{It22} and Matthew correlation coefficient (MCC) \cite{chic, jurm122,and54, It22}.
These indicators require a prior threshold for using them. Thus they all act as \emph{discrete signals }.
In the following sections, we discuss the derived drawbacks of \emph{discrete signals } and suggest solutions.

\section{Discrete Signals}
In this section, we discuss the disadvantages of discrete signals.
To do so, we need to review the typical inference process.

\subsection{ Inference Overview}
Consider a well trained model \textbf{M} and an evaluation set \textbf{D\textsc{test}}
\begin{figure}[hbt!]
\centering
\includegraphics[width=9cm]{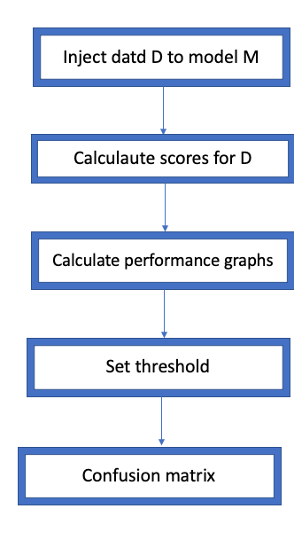}
\caption{ Generic Inference Process}
\end{figure}
one can easily deduce from\textbf{fig 1} that the confusion matrix fully determines the model's evaluation. It leads to the following definition.

\begin{definition}[Discrete signal]  
Let $M$ be the confusion matrix. Consider the function \\ 
   $F   \colon M\to R$ \\
 If $F$ is monotone for each entry of $M$, then F is a \textbf{Discrete signal}. \\  If $F$ does not depend on $M$ then it is called  \textbf{Continuous signals}.
    \end{definition}  
We note that the domain on the Discrete signal can be every nonempty subset of the entries of $M$. 

The output of a classification model is a probabilities vector  \cite{torch,skl}.
We use these vectors to calculate \textbf{FR} and \textbf{TR} curves. For classifying the data, we set a threshold. This threshold determines the confusion matrix. This matrix is the domain of the discrete signals \cite{It22}. Most of the common goodness of fit KPIs are discrete signals,
nonetheless, these signals may suffer from three essential disadvantages:
\begin{itemize}
  \item Unstable concerning the threshold
  \item Difficult for risk calculations 
  \item Absence of good mathematical toolbox
\end{itemize}
In the following subsections, we discuss these disadvantages.
 
\subsection{Instability}
Model's performances have a substantial capital impact. Therefore it is crucial to evaluate our indicators accurately. Setting a fixed threshold on the model graphs  may provide two caveats:

\begin{itemize}
  \item Typical  graphs suffer from steep slopes concerning the thresholds   
  \item Real-world statistics do not always identical to the distribution of the evaluation test 
\end{itemize}  
Academically, these phenomena are seldom studied.
Nonetheless, different distributions and steep slopes often indicate instability. Thus, we find these caveats cardinal in the commercial world.

\begin{figure}[hbt!]
\centering
\includegraphics[width=12cm]{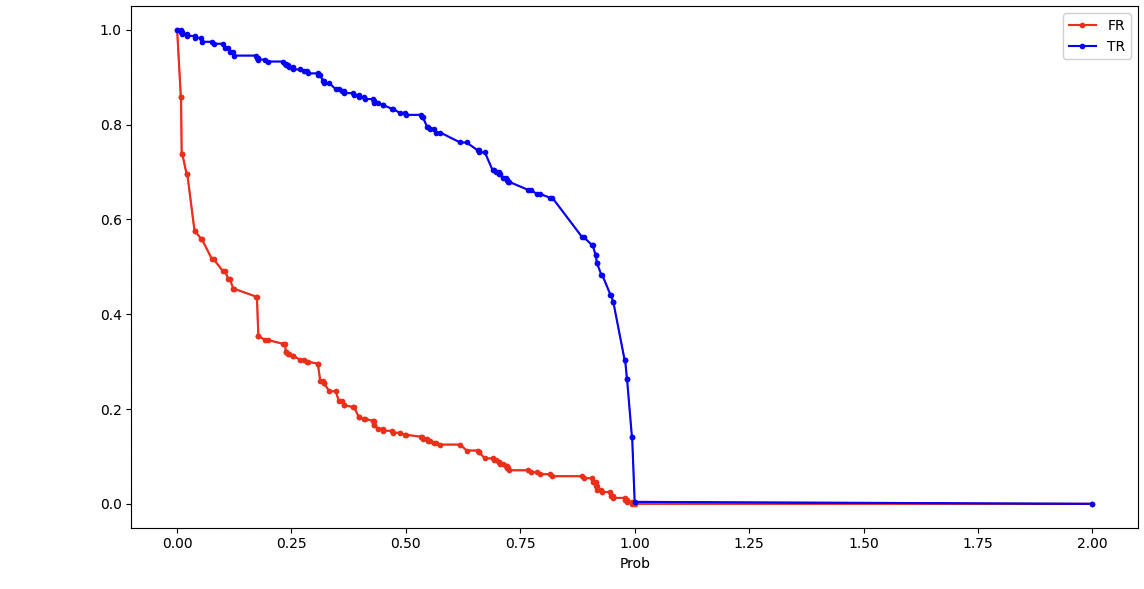}
\caption{ True Accept and FA graphs.  }
\end{figure}

\subsection{Risk Estimation}
A cardinal tool in classical statistics is risk estimation. Whether a statistician is a Bayesian and uses \textbf{credible interval} \cite{CR, RO}  or a frequentist that uses \textbf{confidence interval}  \cite{CON, BF22}, this tool is essential. When we study distribution parameters, the ideal outcome consists of the parameters and a confidence measurement based on the distribution family.  
When we set a threshold or perform statistics such as maximum, we truncate our statistical information and collapse it to a single number.  
We can estimate the risk based on the threshold settings. However, the latter depends on our model, which leads to a non-coherent process.
In contrast to the academy, the model's risk estimation is crucial in the commercial world.

\subsection{Lack of Mathematical Toolbox}
The final disadvantage of discrete signals is motivated by dynamical systems. Since we define discrete signals on the confusion matrix, which is a fixed matrix, we cannot define open sets. We can study neither infinitesimal perturbations nor stability analysis. These two are cardinal for the model's pre-deployment tests.

\subsection{Predict Proba}
The data scientists among the readers  may wonder \textbf{"What about predictproba ?"},\cite{prba}.
Indeed, predictproba is not a discrete signal since it doesn't use a confusion matrix. However, it merely provides a  scores histogram and has no canonical form. Therefore we can have no generic methodology to study its stability or evaluate its risk. Nevertheless, one can consider the discussion in the following sections as \textbf{"Methods for continuous approximation of proba"}

\subsection {So What Can We Do?}
We over-viewed the main drawbacks of discrete signals.  \textbf{Can we provide a remedy?}
If we search for common manners of these drawbacks, it is clear that a more "continuous" framework can be beneficial. Thus defining \textbf{PDFs} on models' curves can assist in this study.

\section{Continuous Signals- Parametric PDF}
\subsection{Motivation}
We discussed the drawbacks of discrete signals. Nonetheless, models output continuous signals: their scores' curves.
If we replace the common analysis that studies a confusion matrix  with an analysis of these curves, we may overcome some of the drawbacks:

\begin{itemize}
\item Curves allows you to calculate different order derivatives which indicate stability status
\item It allows to use of metrics such as 
 the $L^{P}$ or probabilistic such as Jensen-Shannon or KL \cite{KL, JS}. 
\item it allows to obtain the behavior of common indicator upon perturbation
\item Using distribution family manners, it can evaluate risk using the interval of confidence
\end{itemize}
We can cleverly choose a distribution family that handles most of the discrete signals' drawbacks using its parameters. It preserves the probabilistic nature of ML models.

\subsection{Parametric Distributions}
Consider a standard binary classification problem. We train a model using a deep learning architecture or a classical tool such as logistic regression. In the inference, the model outputs a vector of probabilities of length 2 (number of classes).  \textbf{Fig 3} presents a typical scenario.
\begin{figure}[ t!]
\centering
\includegraphics[width=12cm]{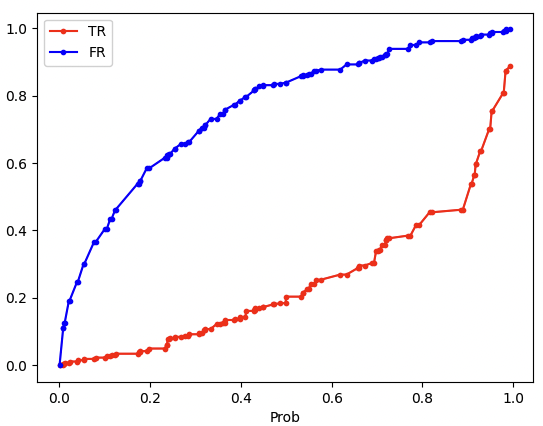}
\caption{ TR (blue) and FR (red)}
\end{figure}
We will give a  mathematical definition that probably most of the readers are familiar with:

\begin{definition}{Cumulative}
A function \textbf{F} is said to be \textbf{a Cumulative Distribution Function} (CDF) if it satisfies the following: 
    \begin{itemize}
        \item{Non decreasing}
        \item{Right continuous}
        \item{$\lim_{x\to -\infty} \textbf{F}(x) =0$}  
        \item{$\lim_{x\to\infty} \textbf{F}(x) =1$}  
    \end{itemize}
\end{definition} \hfill 
\begin{definition}{Density}
We say that a function \textbf{P} is a \textbf{density function} if it is a derivative of a CDF.
\end{definition}
\\ In \textbf{Figure 3}, we can see that \textbf{FR} and \textbf{TR} satisfy the required. If we have an explicit form of the function, we can derive this function,  evaluate risk and perform stability analysis. Moreover, we can calculate error areas analytically, as appears in \textbf{Fig 4}.  
\begin{figure}[ hbt!]
\centering
\includegraphics[width=12cm]{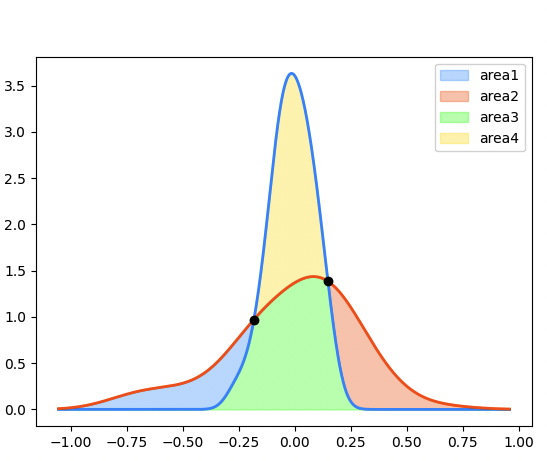}
\caption{ Area 3 represents the intersection between PDFs }
\end{figure}

\subsection{Beta as a Case Study}
Consider a binary classification problem. The model detects whether an input is an element in class \textbf{"1"} and provides the probability for this event. We wish to model the FR and TR using a sound distribution family. A natural choice is \textbf{Beta} function \cite{Ch16,RO,M17}. 
\subsubsection{Beta's Properties}
We will describe  Beta's main properties:
\begin{itemize}
    \item{Beta's support is on $[0,1]$ interval. Moreover, it is strictly great in the interior of the support.  }
    \item{The Beta distribution is infinitely continuous. }
    \item{The distribution has two positive parameters $\alpha$ and $\beta$. }
\end{itemize}
We denote by  $\mu$ the mean of a random variable and by  $\sigma$ the standard deviation. A random variable $X$ with Beta distribution satisfies the following:

\begin{equation}\label{EQ:mu2b}
   \mu[X]   =  \frac{\alpha}{\alpha+\beta} 
\end{equation}
\begin{equation}\label{EQ:v2b}
Var[X]   =
     \frac{\alpha\beta}{(\alpha+\beta)^2  (\alpha+\beta+1 )}
  \end{equation}
We can revert the formula \cite{bfc}:
\begin{equation}\label{EQ:b2mu}
   \alpha    = \left  
                 (\frac{1-\mu[X]}{\sigma[X]^2}- \frac{1}{\mu[X]}\right )\mu[X]^2 
  \end{equation}
\begin{equation}\label{EQ:b2v}
   \beta    = \left  
                 (\frac{1}{\mu[X]} -1\right )\alpha
  \end{equation}
A typical shape of Beta appears in \textbf{Fig 5}. \\
\begin{figure}[ hbt!]
\centering
\includegraphics[width=12cm]{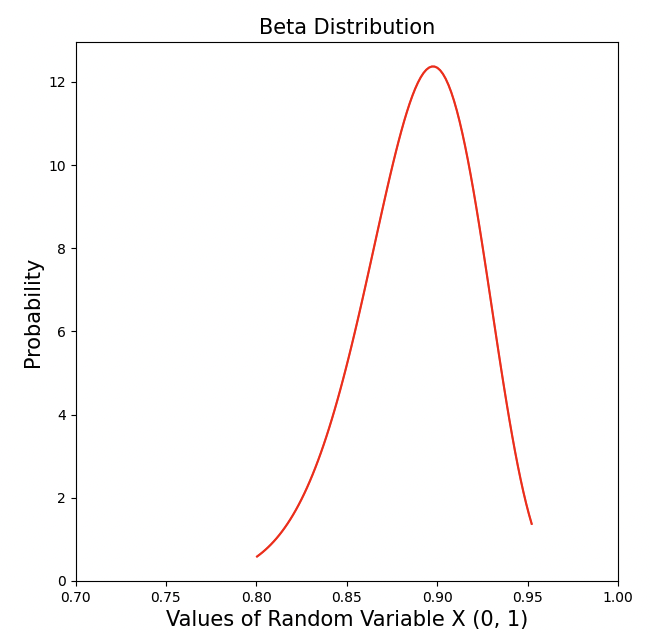}
\caption{ Beta's PDF }
\end{figure}\\
We complete this section by presenting the KL closed form formula of Beta \cite{WIK}. Let positive number $\alpha_{1}$,$\beta_{1}$,$\alpha_{2}$,$\beta_{2}$ 
We have $\Gamma$ and $\Psi$ functions (in some books,$\Psi$ appears as \textbf{digamma} or \textbf{polygamma} or order 0).  
\begin{equation}
\begin{split}
\label{betakl}
               KL[(B(\alpha_{1},\beta_{1})||B(\alpha_{2},\beta_{2})] =   
                \ln\frac{B(\alpha_{2},\beta_{2})}{B(\alpha_{1},\beta_{1})}   +
                 (\alpha_{1}- \alpha_{2})\Psi(\alpha_{1})+\\(\beta_{1}- \beta_{2})\Psi(\beta_{1})
                 +(\alpha_{2}-\alpha_{1} +\beta_{2}-\beta_{1})\Psi(\alpha_{1}+\beta_{1})
\end{split}
\end{equation}

\subsubsection{Example}
In this section, we compare common indicators with the performances of a KL divergence between FR and TR during a model training of a binary classification problem. We follow three indicators
\begin{itemize}
\item {Accuracy} 
\item {MCC}
\item {KL distance between FR and TR}
\end{itemize}

\begin{figure}
\centering
   
   \centering
    \begin{subfigure}{0.5\textwidth}
    \includegraphics[width=0.9\linewidth ]{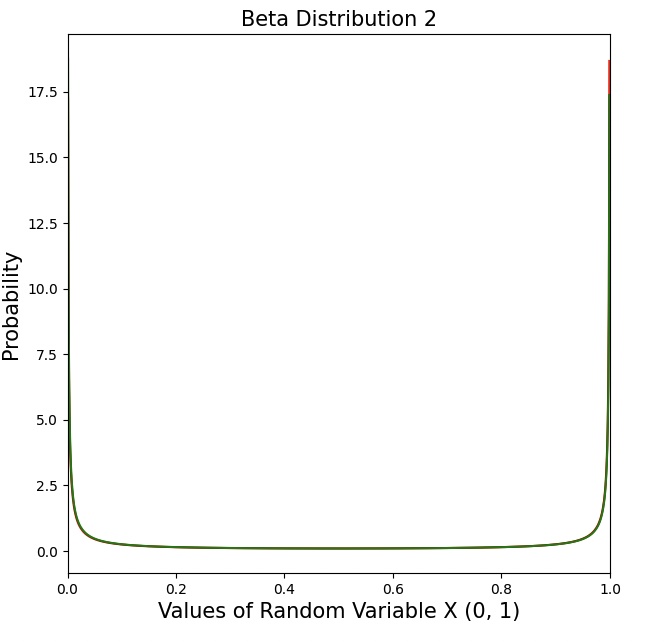} 
    \end{subfigure}
\begin{subfigure}{0.5\textwidth}
\includegraphics[width=0.9\linewidth, height=6cm]{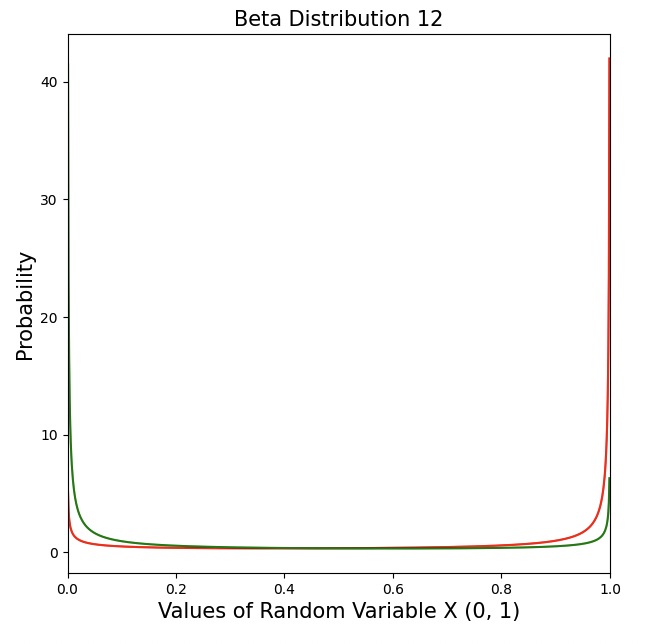}
\end{subfigure}
\begin{subfigure}{0.5\textwidth}
\includegraphics[width=0.9\linewidth, height=6cm]{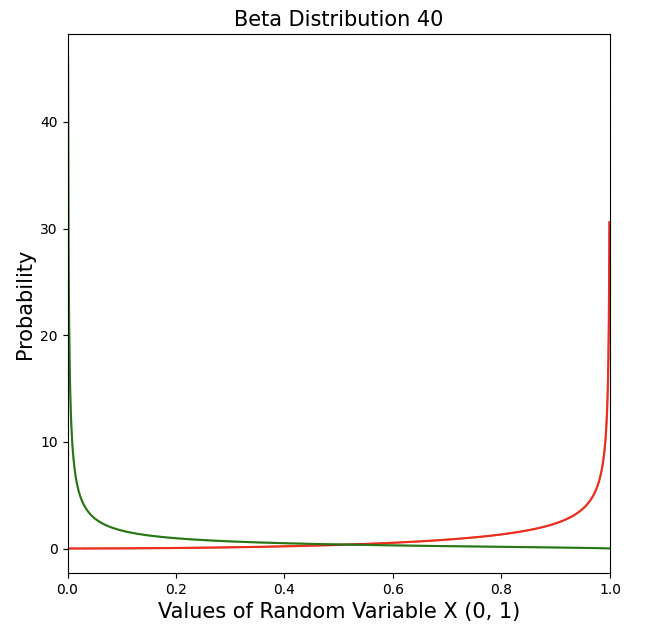}
\end{subfigure}
\caption{Comparison of our measurements using the tuple (accuracy, KL, MCC), As we move down, the models are better trained: Top= ( 0.82,0.04, 0.66), Middle =(0.89,1.06, 0.78), Lowest= (0.91,3.9,0.82).    }
\label{image2}
\end{figure}

The readers can find the graphs in \textbf{Fig 6} and the code is here  \cite{nkg}. The number in the graphs' headers represents the number of epochs. We can see that as this number increases, the gaps between the function increases. More importantly, we see that the KL increases with the accuracy and MCC, which gives an optimistic perspective on our hypothesis.

\subsection{Goodness of Fit - Summary}
We proposed the continuous signal approach and discussed its theoretical improvements for the discrete signals as a goodness-of-fit method. 
We presented a real-world example of this approach for studying a binary classification problem. For modeling the curves, we used Beta distribution and KL divergence.
In the next section, we will study another approach for using continuous signals.

\section{Training}
In the previous sections, we tested the idea that the separation between TR and FR graphs can be a goodness of fit indicator. We have seen some examples that this hypothesis works well. It leads to a further question: Can we use this approach during training by adding a regulation term?
We begin the discussion by presenting an intuition for using this method,\textbf{(It is intuition and not a proof!}).

\begin{definition}[Left epsilon-Beta]
Consider a Beta distribution and a positive small  $\epsilon$. A \textbf{ Left $\epsilon$-Beta function } is a Beta distribution where
\begin{equation} 
\frac{\beta}{\alpha} < \epsilon
\end{equation} 
The right Beta function is defined by the reciprocal  (see \textbf{fig 7}).  
\end{definition}  \\
\begin{figure}[ hbt!]
\centering
\includegraphics[width=12cm]{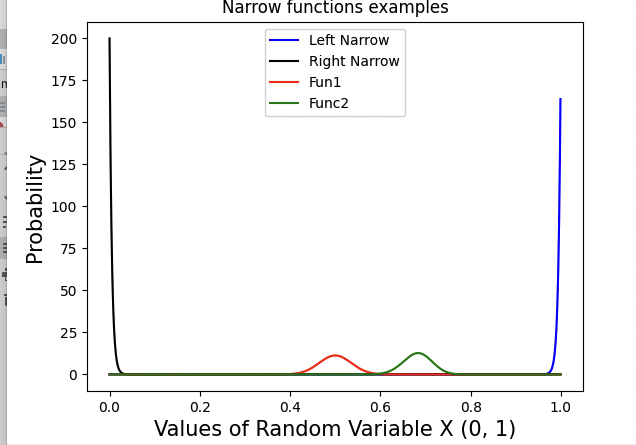}
\caption{ Epsilon Beta functions  }
\end{figure} \\
We aim to maximize the distance between two Beta functions \textbf{P} and  \textbf{Q}. Consider a metric \textbf{d} that satisfies the triangle inequality (KL and J-S do not always do). Let  \textbf{R, L} right and left $\epsilon$-Betas.
The following inequalities hold:
\begin{equation}
\label{ineq} 
    \textbf{d(R,L)} \geq    \max[\textbf{d(P,Q)}]
\end{equation}

\begin{equation}
\begin{split}
\label{ineq}
              \textbf{d(R,L)} \leq  \min[{ \textbf{d(L,P)}+\textbf{d(P,Q)}+\textbf{d(Q,R)} }]   \leq \\ \min[{ \textbf{d(L,P)}+\textbf{d(Q,R)}}] +\max(\textbf{d(P,Q)})
\end{split}
\end{equation}
The LHS is constant. Combining with the upper inequality, we obtain that for some cases maximizing \textbf{d(P, Q)} is equivalent to minimizing the other terms of the RHS,
\subsection{Training Example}
We present an xgboost model training:    \cite{xg}: one model is a vanilla xgboost, and the other uses a new regulation term: the gradient of KL divergence  \cite{nkg}.
We compared the models using three indicators:
\begin{itemize}
\item Accuracy
\item Precision 
\item MCC
\end{itemize}
The results are in \textbf{Fig 8}. 
\begin{figure}
\centering
\begin{subfigure}{0.5\textwidth }
\includegraphics[width=0.9\linewidth, height=6cm]{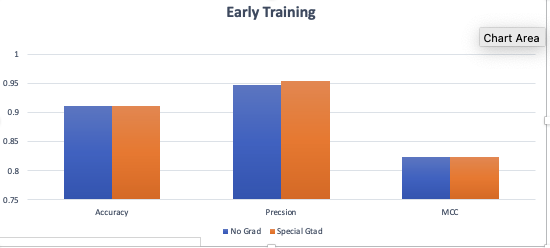} 
\end{subfigure}
\begin{subfigure}{0.5\textwidth}
\includegraphics[width=0.9\linewidth, height=6cm]{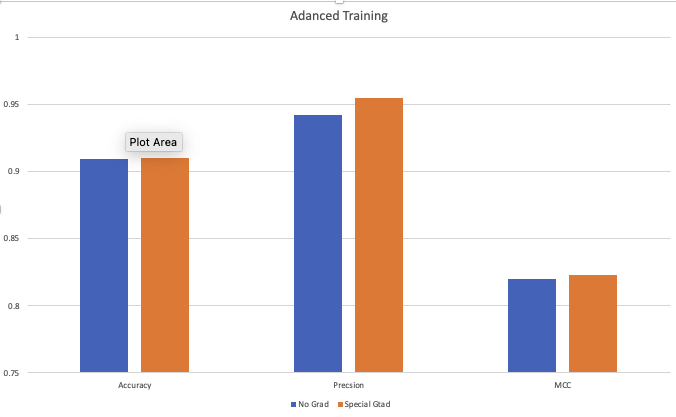}
\end{subfigure}
\caption{Comparison of KPIs between vanilla model and regulation term}
\end{figure}\\
Considering these results, we can't declare a clear winner. However, the "regulated model" shows an advantage in all KPIs compared to the vanilla model. It hints that this approach is not far-fetched and requires further study.

\section{Summary and Future Work}
We described the approaches for evaluating the goodness of fit of ML models and discussed some of their inherent failures. We presented new notions:  \textbf{discrete signals} and \textbf{continuous signals} that allowed us to develop a  different methodology to overcome these failures.
We suggested that parametric PDFs can act as continuous signals and that by using these, we can evaluate the model's risk and analyze its stability. 
We tested this approach for both the goodness of fit purposes and as a training regulation function.
The results are promising, but it is evident that further massive research is required:
\begin{itemize}
\item {Test on various databases} 
\item {Test on different methodologies such as DL}
\item {Generalize binary problems to multi-classes  by replacing Beta to Dirichlet }
\item {Test  \textbf{Isotonic Regression} \cite{Iso} which is extremely common in regression problems }
\item {Test on various distributions such as Gamma}
\end{itemize}
These are all plausible tools for improving the offered approach and enhancing its usage.
Finally, we believe such frameworks will enhance the usage of classical statistics and dynamical system tools.
These tools are mandatory in deploying prediction models, particularly in the commercial world.

\printbibliography 
\end{document}